# On the Choice of Regions for Generalized Belief Propagation

Max Welling (*welling@ics.uci.edu*)
School of Information and Computer Science
University of California Irvine, CA 92697-3425

## Abstract

Generalized belief propagation (GBP) has proven to be a promising technique for approximate inference tasks in AI and machine learning. However, the choice of a good set of clusters to be used in GBP has remained more of an art then a science until this day. This paper proposes a sequential approach to adding new clusters of nodes and their interactions (i.e. "regions") to the approximation. We first review and analyze the recently introduced region graphs and find that three kinds of operations ("split", "merge" and "death") leave the free energy and (under some conditions) the fixed points of GBP invariant. This leads to the notion of "weakly irreducible" regions as the natural candidates to be added to the approximation. Computational complexity of the GBP algorithm is controlled by restricting attention to regions with small "region-width". Combining the above with an efficient (i.e. local in the graph) measure to predict the improved accuracy of GBP leads to the sequential "region pursuit" algorithm for adding new regions bottom-up to the region graph. Experiments show that this algorithm can indeed perform close to optimally.

## 1 INTRODUCTION

Belief propagation (BP) and a growing family of variants thereof have established themselves in recent years as a viable alternative to more traditional approaches for approximate inference in graphical models. BP, like other mean field methods, is likely to exhibit a certain bias in its estimates. However, unlike most sampling schemes, it does not suffer from high variance and is often much more efficient. As a result many exciting applications have emerged in the recent years ranging from vision, operation research, decision making, communication and game theory to learning and error-correcting-decoding.

Despite these favorable properties, BP is no "silver bullet". For highly connected graphs with strong conflicting interactions BP may not produce accurate estimates or may even fail to converge altogether. Convergent alternatives, searching for the same solutions as BP, have been developed but are relatively slow to converge [Yuille, 2002][Welling and Teh, 2001],[Heskes, 2003]. Other shortcomings have resulted in many interesting improvements over the "plain vanilla" BP algorithm. For instance, a convex lower bound for the free energy [Wainwright et al., 2002], "expectation propagation" for continuous valued random variables in the exponential family [Minka, 2001] and the generalized BP (GBP) algorithm that improves the accuracy by including the entropy of larger clusters in the approximation [Yedidia et al., 2002].

Concerning the latter, it has been noted that the choice of clusters of nodes and their interactions (named "regions") in the GBP algorithm is more of "an art than a science". Handpicked regions may form a suboptimal set, leading to inaccuracies or inefficiencies at best. In this paper we will try to formulate an answer to the question: *what is a good set of regions to use in the GBP algorithm?* We propose a sequential process where new regions are selected from a large pool of candidates and added to the approximation.

## 2 GBP AND REGION GRAPHS

Let $\mathcal{V}$ be the set of all variable nodes $i \in \mathcal{V}$ of a factor graph corresponding to the random variables $x \doteq \{x_i\} \doteq [x_1, ..., x_{\mathcal{V}}]$. Similarly, let $\mathcal{A}$ be the set of all factor nodes $a \in \mathcal{A}$ corresponding to potential functions $\{\psi_a\} \doteq [\psi_1, ..., \psi_{\mathcal{A}}]$. A subset of variables $\{x_i \mid i \in \alpha\}$ will be denoted by $x_\alpha$. Factor graphs are bi-partite because the neighbors of a factor, $\mathcal{N}_a$, are precisely all the variables contained in that factor and the neighbors of a variable node, $\mathcal{N}_i$, are precisely all factors that contain that variable. In the following we will often use a shorthand for the variables in the argument of a factor: $x_{\mathcal{N}_a} \doteq x_a$. The expression for the probability distribution corresponding to a factor graph is given by,

$$P_X(X = x) = \frac{1}{Z} \prod_{a \in \mathcal{A}} \psi_a(x_a) \quad (1)$$

where $Z$ is the normalization constant. In the following we will write $P_X(X = x) = p(x)$ to simplify notation.

In [Yedidia et al., 2002] region graphs (RG) were introduced as the natural graphical representation for GBP algorithms (a similar idea was independently proposed in [McEliece and Yildirim, 1998]). Define a region $r_\beta$ as the set of variables $\{i \mid i \in \beta\}$ together with a *subset* of the factors that contain a subset of the variables in $\beta$, $\{a \mid \mathcal{N}_a \subseteq \beta\}$. Thus, although we insist on the fact that the factors in $r_\beta$ can only contain variables in $\beta$, we do not necessarily include all such factors [Yedidia et al., 2002].

A *region graph* is a directed acyclic graph with vertices corresponding to regions. Directed edges can only exist between regions and direct subregions where a direct subregion will be defined as a region that contains all or a subset of the nodes and factors present in the parent region (i.e. it could be a copy of itself). A region $r_\alpha$ is a parent of a region $r_\beta$ if there is a directed edge from $r_\alpha$ to $r_\beta$. Conversely, $r_\beta$ is said to be a child of $r_\alpha$. Ancestors and descendants are also defined in the usual way: a region $r_\alpha$ is said to be an ancestor of a region $r_\beta$ if there exists a directed path from $r_\alpha$ to $r_\beta$. Conversely, $r_\beta$ is said to be a descendant of $r_\alpha$ in that case. "Outer regions" are defined to be the regions with no parents, i.e. without any incoming edges, while "inner regions" are all the remaining regions with one or more parents.

With every region there is associated a counting number $c_{r_\beta}$ defined as follows,

$$c_{r_\beta} = 1 - \sum_{r_\alpha \in Anc(r_\beta)} c_{r_\alpha} \quad (2)$$

where $Anc(r_\beta)$ is the set of all ancestor regions of region $r_\beta$. Given the definition of these counting numbers we can now define the *region graph conditions*: The subgraphs $RG(i)$ and $RG(a)$ consisting of the regions containing a variable $i$ or factor $a$ must be (C1) connected and (C2) satisfy:

$$\sum_{r_\beta \in RG(i)} c_{r_\beta} = 1 \;\; \forall i; \qquad \sum_{r_\beta \in RG(a)} c_{r_\beta} = 1 \;\; \forall a \quad (3)$$

Condition C1 ensures that the marginal distributions defined on regions as computed by GBP are all consistent. Condition C2 for variables makes sure that we do not over-count the degrees of freedom associated with any random variable and guarantees that in the case of vanishing interactions we retain the exact results. Condition C2 for factors ensures that each factor contributes only once to the approximation.

The edges of the RG are associated with the messages of the GBP algorithm. One possible way to execute that algorithm is as follows. Initialize all messages randomly (or equal to a constant, say 1). For a message update between region $r_\alpha$ and a child region $r_\delta$ we first compute the marginal distributions $b_\alpha(x_\alpha)$ and $b_\delta(x_\delta)$ as follows,

$$b_\alpha(x_\alpha) = \frac{1}{Z_\alpha} \prod_{b \in r_\alpha} \psi_b(x_b) \prod_{\substack{r_\gamma \in \mathcal{U}_\alpha \\ r_\beta \in r_\alpha \cup Dec(r_\alpha)}} m_{\gamma\beta}(x_\beta) \quad (4)$$

where $\mathcal{U}_\alpha$ is the set of regions that consist of the parents of $r_\alpha$ and all the parents of the descendants of $r_\alpha$, except for those parents that are contained in $r_\alpha \cup Dec(r_\alpha)$. The same equation is used to compute $b_\delta(x_\delta)$. In terms of these marginals, the message update is then given by the following equation,

$$m_{\alpha\delta}(x_\delta) \leftarrow \frac{\sum_{x_\alpha \setminus x_\delta} b_\alpha(x_\alpha)}{b_\delta(x_\delta)} m_{\alpha\delta}(x_\delta) \quad (5)$$

After convergence, Eq.4 should be used to compute the final estimates of the marginal distributions. We have chosen to present the GBP equations in this particular form because of its compactness and transparency, not because it is an efficient way to implement it. For instance, Eq. 5 makes explicit the fact that the messages enforce the constraint, $\sum_{x_\alpha \setminus x_\delta} b_\alpha(x_\alpha) = b_\delta(x_\delta)$ at convergence, but we need not recompute both marginals $b_\alpha(x_\alpha)$ and $b_\delta(x_\delta)$ at every iteration. Depending on the specific structure of the graph under consideration more efficient update schemes exists in terms of messages only [Yedidia et al., 2002] or in terms of messages and marginals where the latter are stored [Minka, 2001].

From the equations above it becomes apparent that messages and potentials fulfill very similar roles. In fact we may interpret potentials as messages that never change, or alternatively we may interpret messages to be dynamic potentials. Thus, we could view messages into subsets of nodes as effective or induced interactions between the variables in this subset.

Finally, we define the "free energy" of the RG,

$$\mathcal{F}_{RG} = \sum_{r_\beta \in RG} c_{r_\beta} \mathcal{F}_{r_\beta} \quad (6)$$

where the free energy of a region $r_\beta$ can be decomposed in an energy and an entropy term:

$$\mathcal{F}_{r_\beta} = -\sum_{x_\beta} b_\beta(x_\beta) \sum_{c \in r_\beta} \log \psi_c(x_c) \\ + \sum_{x_\beta} b_\beta(x_\beta) \log b_\beta(x_\beta) \quad (7)$$

The RG free energy $\mathcal{F}_{RG}$ is an important object since, as was proven in [Heskes, 2003], it acts as a cost function for GBP algorithms in that the fixed points of GBP are local minima of $\mathcal{F}_{RG}$.

## 3 INVARIANT OPERATIONS

The family of all RGs exhibits a certain redundancy in the sense that different RGs, corresponding to different message passing algorithms, support the same set of solutions (or fixed points) of GBP. One simple example of such redundancy is given below:

**Link-Birth:** *Adding directed edges between regions and descendants (other than children) will not change the RG free energy and the fixed points of GBP.*

The reason for this invariance is that the addition of links to distant descendants will not change any ancestor sets and thus, by Eq.2 none of the counting numbers. This implies that (C1) and (C2) still hold and that the free energy remains invariant (Eq. 6). The results in [Heskes, 2003] then show that the fixed points of GBP will not alter. In the following we present a number of more elaborate operations on the RG that leave the free energy invariant, which will later be related to the fixed points of GBP.

**Proposition 1: Split**
*Consider an outer region $r_\alpha$ and define a partitioning of the variables and factors in the region $r_\alpha$ into three sub-sets: $\{r_{\alpha_1}, r_{\alpha_2}, r_\beta\}$ s.t. the variables in $r_\beta$ separate $\alpha_1$ from $\alpha_2$ in $r_\alpha$ (i.e. conditioning on $\beta$ makes $\alpha_1$ independent of $\alpha_2$). Assume furthermore that we choose $r_\beta$ large enough so that the variable and factor nodes of all child regions $\{r_\gamma\}$ of $r_\alpha$ are contained in either $r_{\alpha_1 \cup \beta}$ or $r_{\alpha_2 \cup \beta}$. Then, the RG free energy does not change under the following operation on the RG: Split the region $r_\alpha$ into two subregions $r_{\alpha_1 \cup \beta}$ and $r_{\alpha_2 \cup \beta}$ and a new child region $r_\beta$ and remove all edges emanating from $r_\alpha$. Connect $r_\beta$ to all its direct subregions in $Dec(r_\alpha)$ (including copies of itself). Connect the new regions $r_{\alpha_1 \cup \beta}$ and $r_{\alpha_2 \cup \beta}$ to the remaining child-nodes $\{r_\gamma\}$ that did not receive connections from $r_\beta$. If the new regions $r_{\alpha_1 \cup \beta}, r_{\alpha_2 \cup \beta}$ and/or $r_\beta$ become subregions of other existing regions, they will* not *become their children (i.e. we will not draw new directed edges between them).*

**Proof:** Condition C1 still holds since the only way it could be violated is if a node or factor in $r_\alpha$ would be present in both of the new outer regions $r_{\alpha_1 \cup \beta}$ and $r_{\alpha_2 \cup \beta}$, but not in $r_\beta$. Since $r_{\alpha_1} \cap r_{\alpha_2} = \varnothing$ this is not possible. Next we show that the free energy of the outer region $r_\alpha$ remains unchanged after the split. Key is the fact that $\beta$ separates the clusters $\alpha_1$ and $\alpha_2$. This implies that we can decompose the belief in $\alpha$ as: $p_\alpha = p_{\alpha_1 \cup \beta} \ p_{\alpha_2 \cup \beta} \ / \ p_\beta$. This results in: $\mathcal{F}_{r_\alpha} = \mathcal{F}_{r_{\alpha_1 \cup \beta}} + \mathcal{F}_{r_{\alpha_2 \cup \beta}} - \mathcal{F}_{r_\beta}$ which is the expression for the free energy of the new regions. Finally we need to show that the counting numbers for all other regions remain unchanged. First we note that the ancestor set for non-descendants of region $r_\alpha$ do not change (recall that $r_{\alpha_1 \cup \beta}$ and $r_{\alpha_2 \cup \beta}$ are connected to a subset of the children of $r_\alpha$ while $r_\beta$ is only connected to direct sub-regions in $Dec(r_\alpha)$). For the regions in $Dec(r_\alpha)$ we observe that they become descendants of $r_{\alpha_1 \cup \beta}$ *or* of $r_{\alpha_2 \cup \beta}$ *or* of $\{r_{\alpha_1 \cup \beta}, r_{\alpha_2 \cup \beta}, r_\beta\}$. This is true since (I) all children of $r_\alpha$ are connected to at least one of the three new regions, (II) being a descendant of $r_\beta$ implies being a descendant of $r_{\alpha_1 \cup \beta}$ and $r_{\alpha_2 \cup \beta}$ as well, since they are the are the parents of $r_\beta$, and (III) regions that become descendants of both $r_{\alpha_1 \cup \beta}$ and $r_{\alpha_2 \cup \beta}$ must be contained in $r_\beta$ and must also be descendants of $r_\beta$, since $r_\beta$ was connected to all direct sub-regions in $Dec(r_\alpha)$. These facts and the fact that counting numbers depend only on the total counting number of a region's ancestors (Eq.2) prove the claim. □

The reason that the region $r_\beta$ is potentially larger than the smallest "conditioning-set" $r_\delta$ is because of the possibility that some child regions $r_\gamma$ of $r_\alpha$ may cease to be subregions of either $r_{\alpha_1 \cup \delta}$ or $r_{\alpha_2 \cup \delta}$, resulting in a change of the expression for the entropy. We thus need to choose $r_\beta$ large enough to ensure this fact.

We cannot expect regions with incoming arrows (inner regions) to be reducible because there will always be a message that is a joint function of the nodes in that region.

**Example:** Assume that an outer region is the union of 2 disjoint sub-regions, i.e. the intersection is empty: $r_\beta = \varnothing$. In that case we can simply split the region in two pieces without changing the fixed points of GBP. For example, it is pointless to put 2 variables in an outer region if there is no factor accompanying them.

**Proposition 2: Merge**
*Consider two identical regions (or copies) $r_{\beta_1}$ and $r_{\beta_2}$ with $r_{\beta_1}$ a parent of $r_{\beta_2}$. Assume that the descendants of $r_{\beta_1}$ other than $r_{\beta_2}$ are contained in the descendants of $r_{\beta_2}$: $Dec(r_{\beta_1}) \setminus r_{\beta_2} \subseteq Dec(r_{\beta_2})$. Then, the region based free energy does not change under the following operation on the RG:
Merge the two regions $r_{\beta_1}$ and $r_{\beta_2}$ into one region $r_\beta$*

*and redirect all incoming and outgoing edges from $r_{\beta_1}$ and $r_{\beta_2}$ to $r_\beta$.*

**Proof:** We first note that condition C1 cannot be violated because regions are merged. Next we check condition C2. Define $Anc'(r_{\beta_2})$ to be the ancestors of region $r_{\beta_2}$, excluding $r_{\beta_1} \cup Anc(r_{\beta_1})$. The free energy contributions of $Anc(r_{\beta_1})$ and $Anc'(r_{\beta_2})$ do not change because the rule Eq.2 to assign counting numbers to regions does not depend on regions downstream. The free energy of the regions $r_{\beta_1}$ and $r_{\beta_2}$ before the merge is given by:

$$c_{r_{\beta_1}} \mathcal{F}_{r_{\beta_1}} + c_{r_{\beta_2}} \mathcal{F}_{r_{\beta_2}}$$
$$= (c_{r_{\beta_1}} + (1 - c_{r_{\beta_1}} - c_{Anc(r_{\beta_1}) \cup Anc'(r_{\beta_2})})) \mathcal{F}_{r_\beta}$$
$$= (1 - c_{Anc(r_{\beta_1}) \cup Anc'(r_{\beta_2})}) \mathcal{F}_{r_\beta}$$

where we defined $c_{Anc(r_\alpha)}$ to be the total counting number for the ancestors of a region $r_\alpha$ and $\mathcal{F}_{r_\beta}$ to be equal to $\mathcal{F}_{r_{\beta_1}}$ and $\mathcal{F}_{r_{\beta_2}}$. The above expression is clearly equal to the free energy of the new merged region. Finally, we need to show that the same is true for the descendants $r_{\beta_2}$ (recall that $Dec(r_{\beta_1}) \setminus r_{\beta_2} \subseteq Dec(r_{\beta_2})$). Using Eq.2 we note that the counting numbers only depend on the total counting number upstream of that region. Above we have shown that the total counting number of the regions $r_{\beta_1}$ and $r_{\beta_2}$ is the same as the total counting number of the merged region (and the counting numbers of the ancestors don't change as well). This together with the fact that the ancestor set doesn't change proves the claim. □

This merge operation allows us to simplify the graph that results from a split operation described above. For instance, if the new child region (say $r_{\beta_1}$) already exists as a child region of $r_\alpha$ (say $r_{\beta_2}$), and if $Dec(r_{\beta_1}) \setminus r_{\beta_2} \subseteq Dec(r_{\beta_2})$ then we can merge the two regions.

**Proposition 3: Death [Yedidia et al., 2002]**
*Consider a region $r_\beta$ with counting number $c_{r_\beta} = 0$. Then, the RG free energy does not change under the following operation on the RG:*
*Remove $r_\beta$ and all its incoming and outgoing edges from the RG. Connect all parents of $r_\beta$ to all children of $r_\beta$.*

**Proof:** Condition C1 is not violated since all ancestors and descendants of region $r_\beta$ remain connected after the removal of $r_\beta$. Counting numbers of the ancestors of $r_\beta$ will not change because of Eq.2. The free energy of $r_\beta$ itself does not change because $c_{r_\beta} = 0$ before the removal. The counting numbers of the descendants only depend on the total counting number of their ancestors. However, the only change to the ancestors of $Dec(r_\beta)$ is that region $r_\beta$ has been removed which had counting number $c_{r_\beta} = 0$, implying that all counting numbers of $Dec(r_\beta)$ remain unchanged and thus that

the free energy remains unchanged. □

**Corollary: Invariance of GBP**
*Consider $RG_1$ and a sequence of Split, Merge and/or Death operations resulting in $RG_2$. If both $RG_1$ and $RG_2$ do not contain copies of regions or regions with counting number $c = 0$, the fixed points of the GBP algorithm corresponding to those regions graphs will have the same fixed points.*

**Proof:** The proof is a simple result of the claim in [Yedidia et al., 2002] that the fixed points of a GBP algorithm corresponding to a RG are the stationary points of the corresponding RG free energy.

The proof in [Yedidia et al., 2002] connecting fixed points of GBP with stationary points of $\mathcal{F}_{RG}$ is only valid for RGs with no copies and without regions with $c = 0$. However, it was also noted in [Yedidia et al., 2002] that the GBP algorithm is well defined in the presence of $c = 0$ regions. The same is true for RGs with copies, but we avoid making claims for these cases because of the lack of proof. There are however alternative algorithms that directly minimize the RG free energy [Yuille, 2002][Welling and Teh, 2001][Heskes, 2003]. For these algorithms we can clearly claim that the solution space (set of local minima) is unaltered by any split, merge and death moves introduced above.

**Definition:** *An outer region of a RG is defined to be "weakly irreducible" if it cannot be split into smaller regions by a sequence of split, merge and death moves without introducing new child regions. When a region cannot be split into smaller pieces, even if the introduction of new child regions is allowed, we will call it "strongly irreducible".*

**Example:** Consider an outer region consisting of 4 nodes and 4 edges that form a cycle, where the edges denote pairwise interactions. It's children are 4 regions containing pairs of nodes and one interaction (or factor) each. This region is clearly weakly irreducible. However, we may split the cycle into two triangles and add the intersection (a chord) as their child. Thus, this cycle is reducible in the strong sense.

By applying the 3 operations repeatedly to outer regions we may be able to simplify the original RG and improve computational efficiency. As an extreme case of this, we may define one region that contains all factors and variables. Any RG resulting from a sequence of split, merge and death operations (with no copies and $c = 0$ regions) will correspond to an *exact* GBP algorithm. For example, decomposable models can be reduced to two layer RGs corresponding to junction trees with cliques as outer regions and separators in the second layer. As a special case we have that trees can be decomposed into two layer regions graphs with edges as outer regions and variables as their child regions.

## 4 COMPLEXITY ISSUES

The conclusion from the previous section is that different outer regions in a RG may nevertheless result in exactly the same fixed points for GBP. This allows us to somewhat organize our search for promising candidate regions to be added to the RG, and limit consideration to simple building blocks, which we will take to be the *weakly irreducible* regions. Although we acknowledge that this is not the only choice there are a number of reasons we have chosen it. Allowing a further reduction of weakly irreducible (but *strongly reducible* regions) into smaller building blocks may cause certain difficulties. For instance, we may introduce copies for which we have no guarantees for an accompanying GBP algorithm (note that $c = 0$ regions can easily be cleaned up using death operations). Also, if we decompose a region into subregions we have to triangulate it in such as way that the children remain sub-regions of some region. There are many ways to triangulate which would result in a messy protocol. Unfortunately, even the set of weakly irreducible building blocks leaves us with a very large number of candidates to consider and there is a need to further organize this search. This will be achieved by taking the complexity of the corresponding GBP algorithm into consideration.

From Eq.5 we see that the bottleneck calculation in GBP is the marginalization operation: $\sum_{x_\alpha \setminus x_\delta} b_\alpha(x_\alpha)$. Thus, the resulting messages are joint functions over the variables in the child regions which implies that the corresponding GBP algorithm scales exponentially in this number. However, also note that often we can exploit some structure in the outer regions to compute the new messages and we don't necessarily have to represent the full joint distributions over the variables in the outer regions. This leads us to define the width of a region:

**Definition:** *The region-width $\omega$ of a region $r_\alpha$ is the tree-width of the induced graph where every variable in $\alpha$ is a vertex and where the variables in the arguments of factors and the variables in the arguments of child regions form cliques (i.e. the variables in factors and child-regions are fully connected).*

**Corollary:** *The complexity of sending messages from a region $r_\alpha$ to all its child regions scales as $D^{\omega+1}$ where $D$ is the number of states per variable.*

**Proof:** In the proof we assume that a child region $r_\delta$ has at least two parents since otherwise its counting number vanishes ($c = 0$) implying that it can be removed using a death move. From Eq.5 we see that the computational complexity of sending a message is dominated by marginalizing $b_\alpha(x_\alpha)$ to $b_\delta(x_\delta)$. From Eq.4 we see that this problem is equivalent to an inference problem where factors *and* incoming messages form cliques in the induced graph. Incoming messages are joint functions of the variables in the child regions $r_\delta$ and subsets thereof (corresponding to other descendants). The computational complexity of this inference problem is precisely governed by the tree-width of the induced graph as defined above. □

These observations suggest that we should specify the computational complexity that we are willing to spend in terms of the maximal allowed region-width before hand. A general procedure for determining all weakly irreducible regions with region-width smaller or equal to a pre-specified value is probably intractable. Instead one could use a heuristic to select promising weakly irreducible candidate regions and remove all regions from consideration that have too large region-width[1].

## 5 ADDING OUTER REGIONS

In this paper we propose to build RGs sequentially and "bottom-up' instead of the usual "top-down" strategies described in [Yedidia et al., 2002], [McEliece and Yildirim, 1998]. Since incorporating new regions will change the approximation we need to make sure that the resulting RG is still valid. The following lemma will set the conditions for that to be true:

**Lemma:** *Consider a valid RG and its sub-graphs $RG(a)$ $\forall a$ that consist of all regions containing factor $a$ and $RG(i)$ $\forall i$ that consist of all regions containing variable $i$. If for all factors $a$ and all variables $i$ the sub-graphs $RG(a)$ and $RG(i)$ contain exactly one leaf node, then adding a new outer region to RG by connecting it to all its direct sub-regions and recomputing the counting numbers according to Eq.2 will result in a new valid RG.*

**Proof:** Since a child must be a sub-region of a parent, the sub-graph $RG(a) \setminus r_{\mathbf{leaf}(a)}$ must be the ancestor set of $r_{\mathbf{leaf}(a)}$. The counting number for $r_{\mathbf{leaf}(a)}$ is exactly equal to $c_{r_{\mathbf{leaf}(a)}} = 1 - \sum_{r_\alpha \in Anc(r_{\mathbf{leaf}(a)})} c_{r_\alpha}$ which implies that the total counting number is precisely 1. The same reasoning is valid for $RG(i)$. The above implies that C2 still holds true and it is trivial to check that C1 is also still true. □

RGs that remain valid if we add new outer regions will be called *"extendable"* hereafter. The RG corresponding to the Bethe approximation is the simplest extendable RG. This fact suggests to use the Bethe approximation as our base approximation on which to build more complex RGs by adding new outer regions.

---

[1]There are efficient algorithms to check the tree-width of a given graph [Bodlaender, 1997].

Next we assume that using the considerations of the previous sections we have a list of candidate regions to be evaluated for their inclusion into the RG. Since we don't have access to the exact marginal distributions we need to find an approximate measure that predicts the improvement in our approximation after including the new candidate. Moreover, this measure should be "cheap" to evaluate since we need to compute it for every candidate region. To make progress we will first make the following assumption:

**Assumption 1:** *The approximation will improve if we add a new region to the RG.*

This assumption is in accord with our intuition that treating larger clusters of nodes jointly in the approximation will improve the accuracy at a cost of increased computational complexity. However, it is by no means true under all circumstances. For instance, in the regime of very strong interactions adding larger clusters may deteriorate the approximation rather then improve it. Also, when the counting numbers become large, as may happen in densely connected graphs, the approximation is typically poor and there are no guarantees that adding regions will be helpful. Thus, the assumption that we make here is that we operate in a regime where GBP is expected to give good results, which can be further improved by adding more and larger regions.

Under the above assumption we want to add the region which induces the maximal *change* in the marginal distributions or the RG free energy which we assume are closely correlated. In figure 1 we compare absolute changes in marginals, free energy and entropy for a loop of size 5 with varying strengths of interactions and incoming messages. In order to evaluate the proposed measure, we still require to iterate GBP to convergence for each candidate region that we wish to consider. This is clearly quite expensive computationally which leads us to the following further assumption:

**Assumption 2:** *The change in the free energy contribution after adding a new region to the RG is relatively small for all non-descendants of the new region.*

The intuition behind this assumption is that the counting numbers of the descendants of the new region will change, while the counting numbers of the remaining nodes remain fixed. Their free energy contribution is only affected by the change in their marginal distribution. Another way of seeing this is to assume for a mo-ment that the all messages in GBP have converged. By entering the new region we need to compute new messages from this new region into its children which send new messages to their children etc. until all leafs of the RG have been reached. At that point the marginal distributions of the descendants have become consistent with that of the new region. In the next stage of GBP we now propagate these changes into the rest of the graph which will then have a "back-reaction" to the new region and its descendants etc. What we propose is thus to ignore this "back-reaction" and keep all messages that flow into the new region and its descendants frozen when we compute the change in the free energy. This has the important effect that this change can be computed *locally* in the graph and will thus save us computation. In the following, we will denote this local change in the free energy by $\Delta^\ell \mathcal{F}$. In the experiments we will verify that this approximation is reasonable.

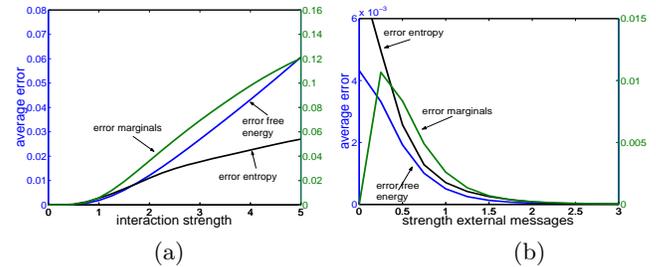

Figure 1: Impact of changing the interaction strength (a) and the strength of external messages (b) on the error in the free energy & entropy (left axes) and marginal distributions (in $L_1$ norm - right axes). The region is a loop of size 5 with binary states and with interactions on the edges sampled in the log-domain from a Gaussian with std. between $[0, 5]$ in (a) and std. 1 in (b). Messages into nodes were also randomly sampled in the log-domain from a Gaussian with std. 1 in (a) and std. between $[0, 3]$ in (b). Results are averaged over 100 random instantiations. The error in the entropy and in the free energy correlates well with that in the marginals except in (b) at the origin. This vanishing error is "accidental" due to the symmetry that flipping all states has equal free energy implying that $p_i(x_i) = 0.5 \; \forall i$.

## 5.1 A REGION PURSUIT ALGORITHM

We now combine our considerations from the previous sections into a region pursuit algorithm (see table below). As an example we will consider a factor graph with only pairwise interactions and take the Bethe free energy as our base approximation. Thus, our starting point is a RG with factors and pairs of variables in the top layer and single variables in the bottom layer. For new candidate regions we can use the results from section 3 to conclude that we can strip away any tree-like structures that are attached with a single node

| Region Pursuit |
| --- |
| 1  *Choose:* |
| 1a   $W$: the maximal allowed region-width. |
| 1b   $K$: the maximal number of regions to be added. |
| 1c   $k$: the number of regions to be added per iteration. |
| 2  *Preprocessing:* |
| 2a  Run GBP on an extendable base approximation (typically Bethe approximation). |
| 2b  Using some suitable heuristic, generate a large number of candidate regions. Remove all candidate regions with a region-width larger than $W$. |
| 3  *Repeat until $K$ regions have been added:* |
| 3a   For all remaining candidate regions do: |
| 3a-i   Decompose the candidate region into weakly irreducible components and consider these components separately. |
| 3a-ii   Check if the region-width is still smaller than or equal to $W$ (it may have changed). |
| 3a-iii   Add the candidate region as a new outer region and connect it to the all its direct sub-regions. Compute the local change in the region free energy $\Delta^\ell \mathcal{F}$ (using fixed messages from GBP). |
| 3b  Add the $k$ regions with largest $\Delta^\ell \mathcal{F}$ to the RG. |
| 3c  Re-run GBP to convergence on the enlarged RG (without removing $c = 0$ regions). |

to some region under consideration. This implies that the simplest candidate outer regions must be weakly irreducible with region-width 2. It may still be too difficult to find all weakly irreducible width-2 regions and so we start with the ones which have only 3 variables and interactions among each pair of variables, i.e. a loop of size 3. Next we consider loops of size 4 without chords (otherwise it would be weakly reducible) etc. Adding *loops* as new regions is in accordance with our intuition that they are the primary cause of inaccuracies in the Bethe approximation since evidence can travel around and may be double counted as a result. Hence, adding loops to the approximation seems a natural first step towards improving the Bethe approximation.

## 6  EXPERIMENTS

The following experiments were designed to test the two assumptions presented in section 5 for evaluating new candidate regions. We used factor graph models with pairwise interactions and binary states. In the following $\mathcal{SG}_{n \times m}$ will denote a square grid model of size $n \times m$ while $\mathcal{FC}_n$ will denote a fully con-

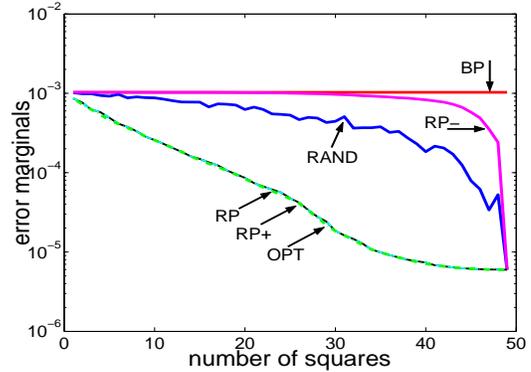

Figure 2: $\mathcal{SG}_{8\times 8}$: Square grid with 64 binary nodes and $W_{max} = 1$ and $\alpha_{max} = 0.5$.

nected model with $n$ nodes. Node and edge potentials were generated as "weights" $\{W_{ij}\}$ and $\{\alpha_i\}$ in the log-domain[2] between $[0, \alpha_{max}]$ (nodes) and $[0, W_{max}]$ (edges) in such a way that some clusters of nodes have strong interactions while others have weak interactions. In all experiments we sequentially add squares ($\mathcal{SG}$ models) or triangles ($\mathcal{FC}$ models) and compare the change in accuracy of the approximation (measured as the average $L_1$ error of the node marginals) for the following methods:

**OPT:** At each iteration and for each graph containing a different candidate region we compute the single variable marginal distributions $b_i(x_i) \quad \forall i$ by running GBP. These are compared with ground truth and the region with the smallest average $L_1$ error is chosen.

**RP:** At each iteration we use the region pursuit algorithm of section 5.1 to choose a new region.

**RP+:** At each iteration and for each graph containing a different candidate region we run GBP to convergence and use the change in free energy to evaluate its merit. This is like **RP** but without the approximation to compute the change in the free energy.

**RP-:** The same procedure as region pursuit, but now we choose the region which induces the *smallest* change in the free energy.

**RAND:** We pick a new region at random with equal probability. Results are averaged over 10 random draws.

The results for $\mathcal{SG}_{8\times 8}$ and $\mathcal{FC}_7$ are reported in figures 2 and 3 respectively. For the square grid GBP is expected to improve the approximation and hence assumption 1 is expected to hold. In this case we see from figure 2 that **OPT, RP+** and **RP** always pick the same regions implying that **RP** is performing optimally (given the sequential constraint) and that assumption 2 holds. Moreover, we see that for any given number of added regions **RP** significantly outperforms random choices which in turn performs better than **RP-**. For fully connected graphs we do not expect GBP to necessarily improve the accuracy of the approximation and assumption 1 may break down as a result. This phenomenon is indeed observed in figure 3, where around iteration 12 the performance of **OPT** on the one hand and **RP** and **RP+** on the other start to diverge. Note however that at all times the performance of **RP** and **RP+** is indistinguishable implying that assumption 2 is still valid. The fact that **RAND** and even **RP−** perform superior to **OPT** around iteration 30 reveals that in this regime its better to add regions with weak interactions while **OPT** is stuck with strongly interacting regions added earlier in the process. The fact that we can gain about an order of magnitude in accuracy for fully connected graphs is somewhat surprising, but in the absence of a reliable stopping criterium this result is of little practical value.

---
[2]More precisely, $\psi_{ij}(x_i, x_j) \propto \exp(W_{ij} x_i x_j)$ and $\psi_i(x_i) \propto \exp(\alpha_i x_i)$ with $x_i = \pm 1$.

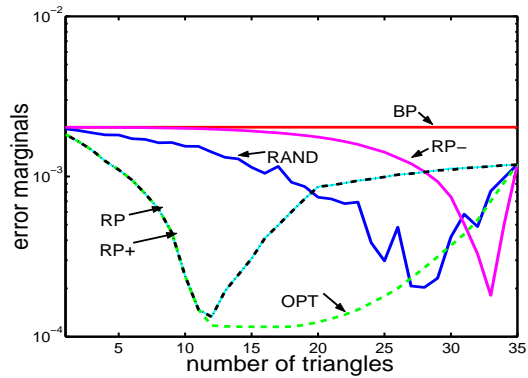

Figure 3: $\mathcal{FC}_7$: Fully connected network with 7 nodes, binary states, $W_{max} = 0.3$ and $\alpha_{max} = 0.5$.

## 7 DISCUSSION

In this paper we have dealt with the choice of regions in GBP. While in highly structured graphical models a good choice of regions may sometimes present itself naturally, most often this is not the case and an automated procedure becomes desirable. We have advocated a sequential approach where new regions are added "bottom-up" to the RG. The only related work we are aware of is on a graph partitioning strategy for the mean field approximation [Xing and Jordan, 2003].

The experiments presented in this paper have been limited to testing the assumptions underlying a novel measure to choose new clusters from a pool of candidates. Future experiments should implement a fully automated procedure to produce promising candidate clusters and evaluate them for their merit. We predict that this will have practical value for problems such as improving the performance of decoders in the field of error-correcting-decoding.

We have observed that convergence of GBP can be somewhat problematic especially in RGs with many generations. Convergent alternatives can replace GBP but they are known to exhibit slow convergence. Improved damping schemes may help elevate this problem.

There are interesting extensions of the region pursuit algorithm which are worth considering. In particular, regions with very high region-widths are currently excluded for computational regions, but a method may be conceivable that can treat these regions in some approximation allowing their inclusion.

## References

[Bodlaender, 1997] Bodlaender, H. (1997). Treewidth: algorithmic tecniques and results. In *Proc. 22nd Int. Symp. on Math. Found. of Comp. Sc.*, volume 1295, pages 29–36. Springer-Verlag.

[Heskes, 2003] Heskes, T. (2003). Stable fixed points of loopy belief propagation are minima of the bethe free energy. In *Advances in Neural Information Processing Systems*, volume 15, Vancouver, CA.

[McEliece and Yildirim, 1998] McEliece, R. and Yildirim, M. (1998). Belief propagation on partially ordered sets. In eds. D. Gilliam and Rosenthal, J., editors, *Mathematical Systems Theory in Biology, Communications, Computation, and Finance*.

[Minka, 2001] Minka, T. (2001). Expectation propagation for approximate Bayesian inference. In *Proceedings of the Conference on Uncertainty in Artificial Intelligence*, pages 362–369.

[Wainwright et al., 2002] Wainwright, M., Jaakkola, T., and Willsky, A. (2002). A new class of upper bounds on the log partition function. In *Proceedings of the Conference on Uncertainty in Artificial Intelligence*, Edmonton, CA.

[Welling and Teh, 2001] Welling, M. and Teh, Y. (2001). Belief optimization for binary networks: a stable alternative to loopy belief propagation. In *Proceedings of the Conference on Uncertainty in Artificial Intelligence*, pages 554–561, Seattle, USA.

[Xing and Jordan, 2003] Xing, E. and Jordan, M. (2003). Graph partition strategies for generalized mean field inference. Technical Report CSD-03-1274, Division of Computer Science, University of California, Berkeley.

[Yedidia et al., 2002] Yedidia, J., Freeman, W., and Weiss, Y. (2002). Constructing free energy approximations and generalized belief propagation algorithms. Technical report, MERL. Technical Report TR-2002-35.

[Yuille, 2002] Yuille, A. (2002). CCCP algorithms to minimize the Bethe and Kikuchi free energies: Convergent alternatives to belief propagation. *Neural Computation*, 14(7):1691–1722.